\documentclass[10pt,final,conference,letterpaper]{IEEEtran}
\pdfoutput=1 
\IEEEoverridecommandlockouts
\usepackage{xspace}
\usepackage{pifont}
\usepackage{subcaption}
\usepackage{array}
\usepackage{wrapfig}
\usepackage{tcolorbox}
\usepackage{svg}

\usepackage[normalem]{ulem}
\usepackage{amsmath}
\usepackage[linesnumbered,ruled,vlined]{algorithm2e}

\usepackage{layouts}

\usepackage{pgf}
\usepackage{pgfplots}
\pgfplotsset{compat=1.9}

\graphicspath{{graphs/pgf/}}

\usepackage[binary-units=true]{siunitx}
\usepackage{balance}

\usepackage{hyperref}
\hypersetup{
    colorlinks=true,
    urlcolor=blue
}

\usepackage{cleveref}
\crefformat{section}{\S#2#1#3}

\newcommand{\ie}{\emph{i.e.,}\xspace}
\newcommand{\eg}{\emph{e.g.,}\xspace}

\newcommand{\gmodel}{$W_{global}^{t+1}$}

\newcommand{\tool}{{MeanCache}\xspace}

\usepgfplotslibrary{external}
\usepackage{fancyhdr}
\usepackage{xcolor} 

\def\BibTeX{{\rm B\kern-.05em{\sc i\kern-.025em b}\kern-.08em
    T\kern-.1667em\lower.7ex\hbox{E}\kern-.125emX}}

\makeatletter
\def\ps@IEEEtitlepagestyle{%
    \fancyhf{} 
    \fancyhead[C]{\textcolor{blue}{\textbf{Accepted at 2025 IEEE 39th International Parallel and Distributed Processing Symposium (IPDPS)}}} 
    \fancyfoot[C]{\thepage} 
    \renewcommand{\headrulewidth}{0pt} 
    \renewcommand{\footrulewidth}{0pt} 
}
\makeatother

\begin{document}

\title{MeanCache: User-Centric Semantic Caching for LLM Web Services}

\author{Waris Gill$^1$, Mohamed Elidrisi$^2$, Pallavi Kalapatapu$^2$, Ammar Ahmed$^4$, Ali Anwar$^4$, Muhammad Ali Gulzar$^1$ \\
$^1$Virginia Tech \quad $^2$Cisco\quad $^3$University of Minnesota \\
 waris@vt.edu, \{melidris, pkalapat\}@cisco.com, \{ahme0599, aanwar\}@umn.edu, gulzar@cs.vt.edu

 }


\maketitle

\pagestyle{fancy}
\fancyhf{} 
\fancyfoot[C]{\thepage} 
\renewcommand{\headrulewidth}{0pt} 
\renewcommand{\footrulewidth}{0pt} 


\begin{abstract}
    Large Language Models (LLMs) like ChatGPT and Llama have revolutionized natural language processing and search engine dynamics. However, these models incur exceptionally high computational costs. For instance, GPT-3 consists of 175 billion parameters, where inference demands billions of floating-point operations. Caching is a natural solution to reduce LLM inference costs on repeated queries. However, existing caching methods are incapable of finding semantic similarities among LLM queries nor do they operate effectively on contextual queries, leading to unacceptable \emph{false} hit-and-miss rates.

This paper introduces \tool, a user-centric semantic cache for LLM-based services that identifies semantically similar queries to determine cache hit or miss. Using \tool, the response to a user's semantically similar query can be retrieved from a local cache rather than re-querying the LLM, thus reducing costs, service provider load, and environmental impact. 
\tool leverages Federated Learning (FL) to collaboratively train a query similarity model without violating user privacy. By placing a local cache in each user's device and using FL, \tool reduces the latency, costs, and enhances model performance, resulting in lower false-hit rates. \tool also encodes context chains for every cached query, offering a simple yet highly effective mechanism to discern contextual query responses from standalone queries. Our experiments benchmarked against the state-of-the-art caching method reveal that \tool attains an approximately 17\% higher F-score and a 20\% increase in precision during semantic cache hit-and-miss decisions while performing even better on contextual queries. It also reduces the storage requirement by 83\% and accelerates semantic cache hit-and-miss decisions by 11\%.

\end{abstract}

\section{Introduction}
Large Language Models (LLMs) like ChatGPT~\cite{chatgpt}, Google Bard~\cite{bard}, Claude~\cite{claude}, and Llama~\cite{llama} have demonstrated remarkable capabilities in understanding and generating human language, leading to significant advancements in applications ranging from search engines to conversational agents. LLMs are increasingly integrated into platforms like the Perplexity AI search engine, 
Rabbit OS~\cite{rabbitr68:online}, and Arc browser~\cite{arc:verge}.

\noindent{\textbf{Motivation.}} 
Generating responses to user queries with LLMs, such as GPT-3, requires substantial computations and poses environmental challenges~\cite{OPTQ, tseng2024quip, sachdeva2024train, ma2024era}. For example, GPT-3's 175B parameters in float16 format consume 326 GB memory, exceeding single GPU capacities and necessitating multi-GPU deployments~\cite{OPTQ}. These requirements lead to high operational costs.  Consequently, LLM-based services charge users and limit query rates~\cite{perplexitypro, openaipricing}. Prior studies have observed that users frequently submit similar queries to web services~\cite{lempel2003predictive, xie2002locality, markatos2001caching} (approximately 33\% of search engine queries being resubmitted~\cite{markatos2001caching}), suggesting opportunities for optimization by avoiding redundant computations.

Caching serves as an effective technique in traditional web services to address duplicate search queries, avoiding redundant computations, significantly improving response time, reducing the load on query processors, and enhancing network bandwidth utilization~\cite{markatos2001caching, lempel2003predictive, podlipnig2003survey, baeza2007impact}. If applicable to LLMs-based web services, such caching can substantially impact billions of floating point operations, thereby decreasing operational costs and environmental impact.

\noindent{\textbf{Problem.}} Existing caching techniques ~\cite{saraiva2001rank, markatos2001caching, xie2002locality, lempel2003predictive} use keyword matching, which often struggles to capture the semantic similarity among similar queries to LLM-based web services, resulting in a significantly low hit rate. For instance, existing caches do not detect the semantic similarity between {\em ``How can I increase the battery life of my smartphone?"} and {\em ``Tips for extending the duration of my phone's power source"}, leading to a cache miss.  Recently, Zhu et al.~\cite{zhu2023towards} and GPTCache~\cite{gptcache} present server-side semantic caching for the LLMs-based services to address the limitations of keyword-matching caching techniques. If a new query is semantically similar to any query in a cache,  the server returns the response from the cache. Otherwise, a model multiplexer selects the most suitable LLM for the query to generate the response.

Existing semantic caches have several limitations. First, they demand a significantly large central cache to store the queries and responses of all users, which is unscalable and violates users' privacy. Second,  they incur the network cost of sending a user query to the server even if there is a cache hit. An end user will still be charged for the query even if the query is served from the server cache. Third,  they use a single server-side embedding model to find the semantic similarity among queries, which does not generalize to each user's querying patterns. 
For instance, Google Keyboard~\cite{gboard} adapts to each user's unique writing style and embeds such personalized behaviors to enhance the accuracy of the next word prediction model. Fourth, they employ Llama-2 to enhance the accuracy of semantic matching~\cite{gptcache}; however, in practice, such models perform billions of operations to generate embeddings, offsetting the benefits of the cache. Lastly, they are only effective on standalone queries, resulting in unbearably high false hit rates for contextually different but semantically similar queries.

\noindent{\textbf{Key Insights and Contributions of \tool.}} This work introduces a novel {\emph {user-centric}} semantic caching system called \tool. \tool provides a privacy-preserving caching system that returns the response to similar queries directly from the user's local cache, bypassing the need to query the LLM-based web service. \tool achieves these goals in the following ways. 

To address privacy concerns associated with central server-side caching, \tool introduces a user-side cache design ensuring that the user's queries and responses are never stored outside of the user's device. To find a semantic match between a new query and cached queries, \tool uses smaller embedding models such as MPNet~\cite{song2020mpnet} to generate embeddings for semantic matching locally. Previous work has shown that a smaller model can achieve performance comparable to larger models on custom tasks~\cite{ouyang2022training, penedo2024refinedweb, du2024compositional}.

Due to different contexts around queries, LLM may return different responses for semantically similar queries. For such queries, \tool also records the contextual chain, parent queries already in the cache, for a given query. To find a response for a contextual query, \tool verifies the context of a contextual query by matching a given query's context with the cached query's context chain to accurately retrieve responses to contextual queries.

Each user may not have sufficient queries to customize an embedding model that can help find a semantic match between new queries and cached queries. To address this, \tool utilizes federated learning (FL), which exploits data silos on user devices for private training for collaborative learning~\cite{10825643, 10386691, mcmahan2017communication, gill2025tracefl, }, thereby personalizing an embedding model for each user. This privacy-preserving training not only customizes the embedding model to the user's querying patterns but also enhances the performance (\ie accuracy) of semantic caching without compromising user privacy (\ie without storing user data on the web server).

The runtime performance of \tool is primarily influenced by the time taken to match a new query embedding vector with existing ones in the cache to find a semantically similar query. The search time is directly proportional to the dimensions of the embedding vector. To optimize runtime performance, \tool compresses the embedding vector by leveraging principal component analysis (PCA)~\cite{pca1, pca2, pca3}, effectively reducing the size of the embedding vector (\ie projecting it to lower dimensional space).  \tool also offers an adaptive cosine similarity threshold, which is also collaboratively computed using FL, to improve accuracy in finding semantic matches between queries.

\noindent{\textbf{Evaluations.}} We compare \tool with GPTCache~\cite{gptcache}. GPTCache is closely related to \tool and has received over 6,000 stars on GitHub~\cite{zillizte18:online}. 
We benchmark \tool's performance against GPTCache on the GPTCache's dataset~\cite{dataset-gptcache} to demonstrate its effectiveness and highlight the improvements \tool offers over existing solutions. 

\tool surpasses GPTCache~\cite{gptcache} by achieving a 17\% higher F-score and approximately a 20\% increase in precision in end-to-end deployment for identifying duplicate queries to LLM-based web services. \tool's performance on contextual queries is even more impressive when compared to GPTCache (baseline). For contextual queries, \tool achieves a 25\% higher F-score and accuracy, and a 32\% higher precision over the baseline. \tool's embedding compression utility approximately reduces storage and memory needs by 83\% and results in 11\% faster semantic matching while still outperforming the state-of-the-art GPTCache. 

\noindent{\textit{\textbf{Artifact Availability:}}} \tool is implemented in the Flower FL framework~\cite{beutel2020flower}. The complete source code and contextual queries dataset will be available at \textcolor{blue}{ \url{https://github.com/SEED-VT/MeanCache}}.

\section{Background}
\label{sec:background}

\noindent{\textbf{Federated Learning (FL).}} FL is a distributed, privacy-preserving ML model training approach~\cite{10.1145/3627703.3650081, 10386691, gill2023FedDebug, gill2023feddefender}. In each FL round, a central server distributes a global model to participating clients, who train it on their local data and return the updated models. The server then aggregates these models to create a new global model for the next round. While several aggregation algorithms exist (FedAvg~\cite{mcmahan2017communication}, FedProx~\cite{li2020federated}, and FedMA~\cite{Wang2020Federated}), FedAvg is most common, using the equation:

\begin{equation}
   W_{global}^{t+1} = \sum_{k=1}^{K} \frac{n_k}{n} w_{k,t}
\end{equation}

where $W_{global}^{t+1}$ represents the new global model, $w_{k,t}$ is the model of the $k^{th}$ client at round $t$, $n_k$ is the sample count at the $k^{th}$ client, and $n$ is the total sample count across participating clients. This process continues for multiple rounds until convergence.




\noindent\textbf{Transformer and Embeddings.} The transformer architecture is based on the attention mechanism~\cite{vaswani2017attention}. Transformers are extensively utilized in natural language processing (NLP) tasks such as machine translation, text summarization, and question-answering. The attention mechanism allows transformers to effectively capture long-range dependencies within sequences, while positional encoding explicitly incorporates information about the order of tokens. Transformers convert text into embeddings, representing words or sentences as dense vectors in high-dimensional space. These embeddings capture semantic meanings, ensuring that semantically similar words or sentences have similar vector representations~\cite{sbert-paper,karpukhin-etal-2020-dense,ni-etal-2022-sentence}. Cosine similarity is a widely used metric for measuring the similarity between embeddings, with values ranging from -1 (completely opposite) to 1 (identical). For instance, embeddings for words such as \textit{cat} and \textit{dog} typically exhibit higher cosine similarity compared to \textit{cat} and \textit{car}, as the former pair is more closely related semantically. For embedding vectors $\mathbf{E}_1$ and $\mathbf{E}_2$, cosine similarity is calculated as:

\begin{equation}
\text{cosine\_similarity} = \frac{\mathbf{E}_1 \cdot \mathbf{E}_2}{||\mathbf{E}_1|| \cdot ||\mathbf{E}_2||}
\end{equation}

where $\mathbf{E}_1 \cdot \mathbf{E}_2$ represents the dot product and $||\mathbf{E}_1||$ and $||\mathbf{E}_2||$ denote the vector magnitudes.


\noindent\textbf{Contextual Queries.} Interaction between end-users and LLM services typically involves two primary types of queries: standalone queries (e.g., Q1 \emph{Draw a line in Python?}) and follow-up, or contextual queries (e.g., Q2 \emph{Change the color to red}). Standalone queries can be resolved independently without any additional information, whereas contextual queries require prior context to provide accurate and relevant responses.

Consider a scenario in which both queries Q1 and Q2, along with their responses, have been cached. If a user later submits a new standalone query, Q3 \emph{Draw a circle?}, followed by a contextual query, Q4 \emph{Change the color to red}, Q4 might appear semantically similar to the cached query Q2. However, despite their similarity, Q4 references a different context (Q3 rather than Q1), thus requiring a distinct interpretation and response. Consequently, a semantic cache without effective context detection may yield incorrect cache hits.

\begin{figure}[t]
    \centering
         \includegraphics[width=0.5\textwidth]{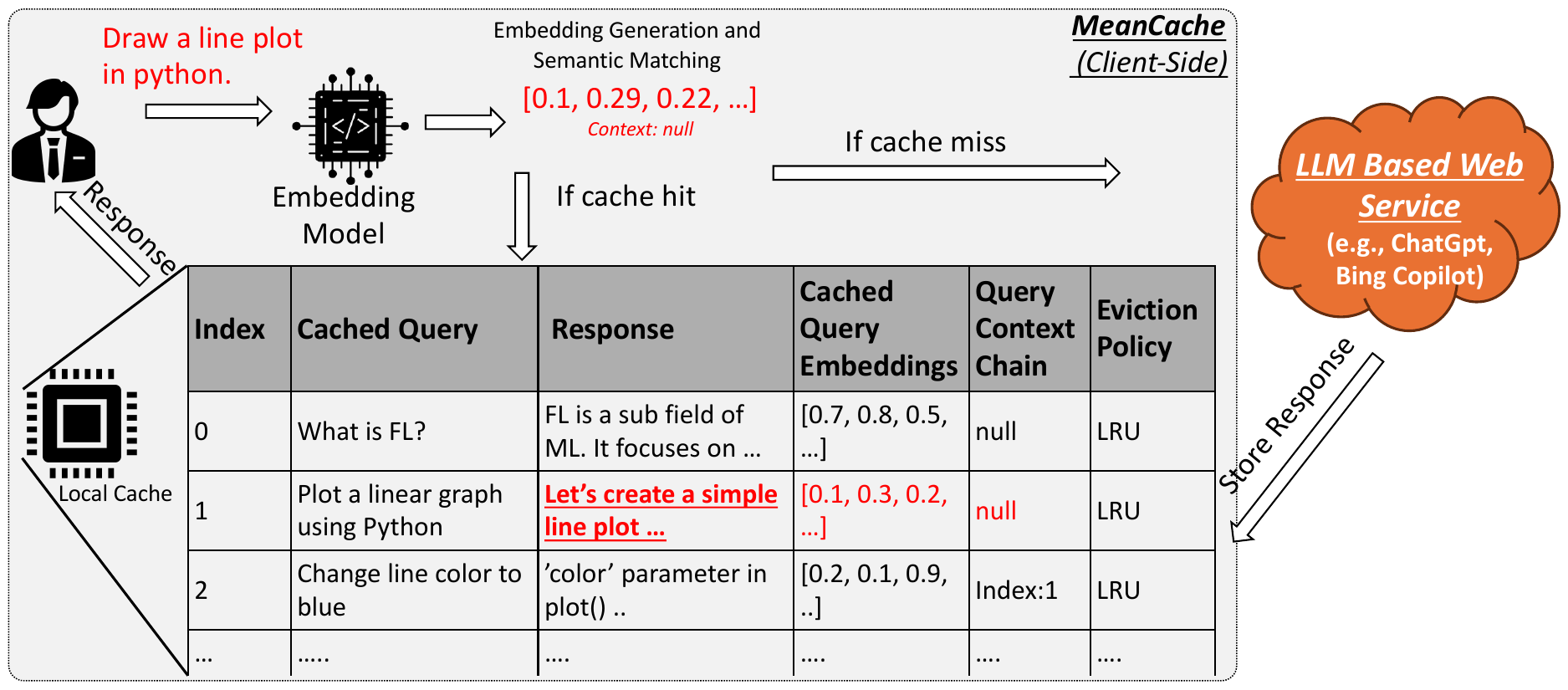}
          \vspace{-2ex}
          \caption{\tool's Workflow.}
          \label{fig:mean-cache-use-case}
          \vspace{-2ex}
  \end{figure}

\section{\tool's Design} 
\label{sec:tool-design}

MeanCache is a user-centric semantic cache optimized for user-side operation. Figure~\ref{fig:mean-cache-use-case} illustrates the workflow of MeanCache for similar queries. Algorithm~\ref{alg:tool} further explains MeanCache's querying and population process programmatically.  When a user submits a query to an LLM web service with MeanCache enabled, MeanCache computes the query embeddings (Line~\ref{line:query-embedding}). These embeddings are then matched with the embedding of the cached queries using cosine similarity (Line~\ref{line:similar-queries}).
For every similar query found within the cache, MeanCache analyzes the context chain for every query and matches it with the conversational history of the submitted query (Line~\ref{line:context-chain}). 
If MeanCache finds a similar query with a similar context chain, the response is retrieved from the local cache and returned to the user (Line~\ref{line:response}). Otherwise, MeanCache forwards the query to the LLM service to obtain the response. The query, its response, and embeddings are then stored in the cache (Line~\ref{line:populate}).

\tool harnesses the collective intelligence of multiple users to train a semantic similarity model, and its user-centric design addresses privacy and scalability issues. To achieve these, \tool takes the following design decisions. It employs a small embedding model with lower computational overhead than LLM based embedding models. It uses FL for collaborative training to fine-tune the embedding model without ever storing user data on a central server. This approach generates high-quality embeddings and improves the accuracy of embedding matching for retrieving similar queries. To handle contextual queries, \tool includes contextual chain information in its cache against every query to identify if the cached response for a query is only applicable under a specific context. This design is capable of handling contextual chains. To reduce storage and memory overhead and expedite search time for finding similar queries in the cache, \tool compresses the embeddings using PCA.

\begin{algorithm}[t]
{\footnotesize
\SetKwInput{KwInput}{Input}      
\SetKwInput{KwOutput}{Output}    
\DontPrintSemicolon

\KwInput{
  User query $Q$, User Query Context $C_q$ (\eg $null$ if no parent)
}
\KwOutput{
  Response $R$
}


    $E_Q \gets \text{encode}(Q)$ \tcp{\scriptsize compute the embedding of the query} \label{line:query-embedding}

    $similar\_queries \gets \text{FindSimilarQueriesinCache}(E_Q)$ 
        \tcp{\scriptsize retrieve top-$k$ similar cached queries} \label{line:similar-queries}

    $context\_match \gets \textbf{False}$ 
        \tcp{\scriptsize flag to indicate if suitable context is found}
    
    \ForEach{context $C_i \in similar\_queries$}{ \label{line:context-chain}

        \If{$C_i$ matches with $C_q$}{
            $context\_match \gets \textbf{True}$\;
            Retrieve response $R$ from cache for $C_i$ \label{line:response}
        }

    }
    
    \If{\textbf{not} $context\_match$}{
        $R \gets \text{LLMResponseAndEnrollInCache}(Q, E_Q, C_q)$
            \tcp{\scriptsize generate response and cache it} \label{line:populate}
            
    }
    
    \KwRet $R$ 
}

\caption{MeanCache}
\label{alg:tool}
\end{algorithm}

\subsection{FL Based Embedding Model Training}
GPTCache~\cite{gptcache} suggests using Llama to generate superior embeddings, thereby enhancing semantic matching accuracy. However, this approach has several limitations. LLMs not only sizable, being gigabytes in size, but they also require substantial computational resources to generate embeddings. Deploying such models, especially at the end-user level, is impractical due to their size and significant computational overhead for semantic matching. \tool employs a compact embedding model, which has a lower computational overhead compared to large embedding models. The smaller model may not provide the same level of accuracy as an LLM. However, a smaller model trained on customized tasks can match the performance of an LLM~\cite{ouyang2022training, penedo2024refinedweb, du2024compositional}. One challenge in using smaller embedding models is that each user may not have sufficient data to train and customize the embedding model. 

\begin{figure}[t]
    \centering
         \includegraphics[width=0.5\textwidth]{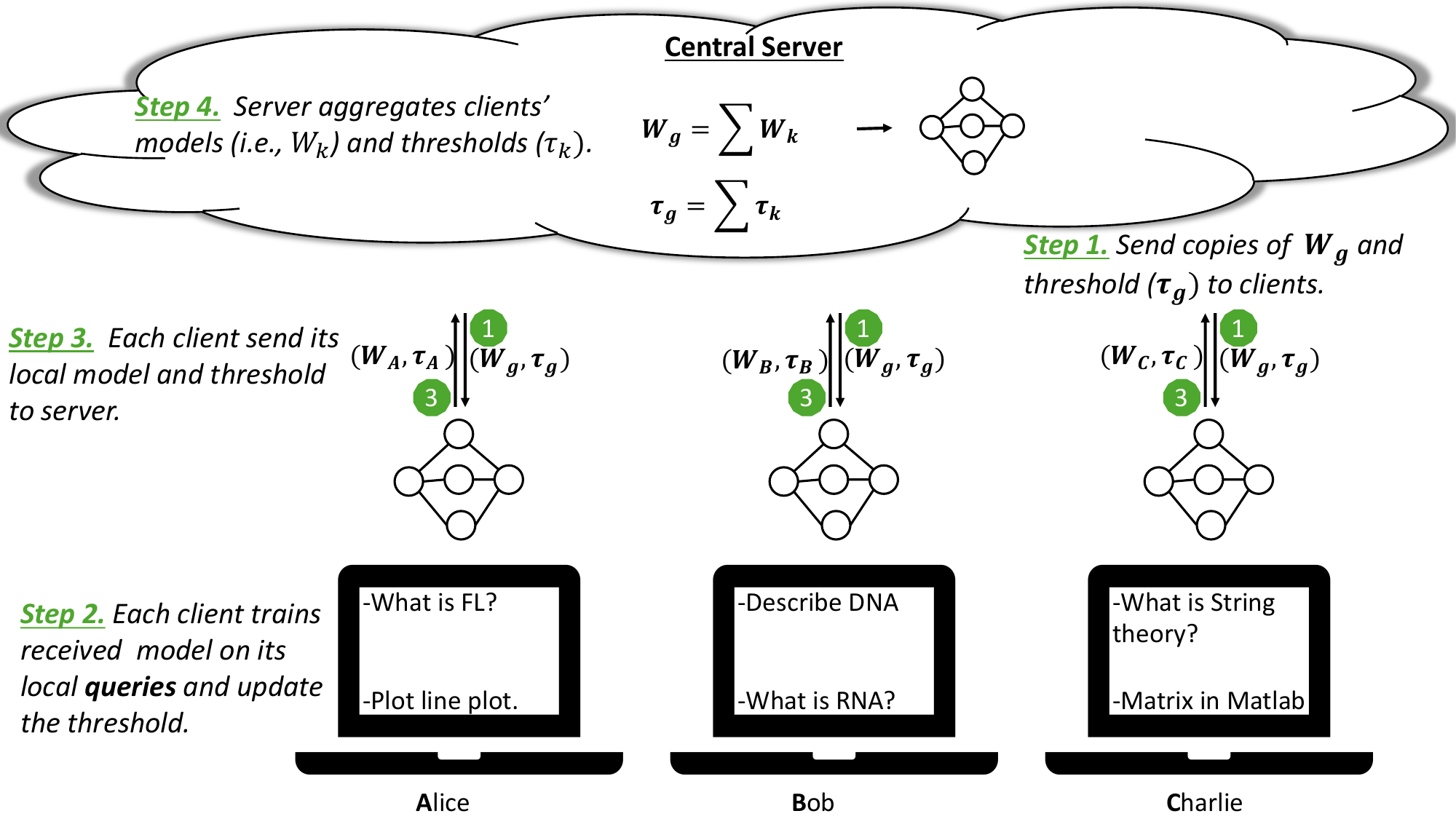}
          \vspace{-2ex}
          \caption{Privacy-preserving model training in \tool.}
          \label{fig:fl-training}
          \vspace{-2ex}
  \end{figure}

\tool utilizes FL to exploit the vast amount of distributed data available on users' devices to train and personalize the smaller embedding model. FL allows the users to train the embedding model locally and learn the optimal threshold for cosine similarity. The updated weights and local threshold are shared with the server. The server aggregates the updated weights and cosine similarity threshold from multiple users to update the global model, which is redistributed back to the users. This approach ensures that the user's privacy is maintained, and the collective intelligence of multiple users is leveraged to improve the performance of the caching system. Figure~\ref{fig:fl-training} shows the overview of privacy-preserving training of the embedding model with FL. 

In the first step, the server sends the initial weights of the embedding model (\gmodel) and global threshold ($\tau$) to a subset of users as shown in step 1 in Figure~\ref{fig:fl-training}. The subset of users is usually randomly selected or selected based on their battery level, network bandwidth, or performance history. Additionally, the server also sends the hyperparameters (\eg learning rate, batch size, epochs) necessary for FL training of the embedding model. 

\subsubsection{Client Training} Upon receiving the embedding model (\gmodel) from the server, each client replaces its local embedding model weights with the newly received weights. Subsequently, each client trains the embedding model locally on its unique local dataset (step 2 in Figure~\ref{fig:fl-training}). To generate high-quality embeddings from unique and similar queries, \tool's clients training employs a multitask learning approach, integrating two distinct loss functions: contrastive loss ~\cite{sbert-paper} and multiple-negatives ranking loss ~\cite{henderson2017efficient, sbert-paper}. These loss functions update the weights of the embedding model during the training process. The contrastive loss function operates by distancing unique (non-duplicate) queries from each other within the embedding space, thereby facilitating the differentiation between duplicate and non-duplicate queries. Unlike contrastive loss, the multiple-negatives ranking loss function minimizes the distance between positive pairs (duplicate queries) amidst a large set of potential candidates \ie multiple-negatives ranking loss does not concentrate on distancing unique queries and its objective is to draw positive pairs (similar queries) closer within the embedding space.

This learning approach enables \tool to adjust to diverse query patterns exhibited by users. For instance, some users may generate more repetitive queries compared to others, while certain users may not produce any repetitive queries at all. Interestingly, \tool's multitask learning objective can benefit from learning even from a user with no repetitive queries. This is because \tool's global embedding model (\gmodel)  will learn to widen the distance between unique queries, thereby effectively learning the true misses of the non-duplicate queries and minimizing the false hits during the search process. True miss happens when a similar query is not present in the cache. A false hit is when a query is found and returned from the cache, which is not actually similar.

\subsubsection{Finding the Optimal Threshold for Cosine Similarity} After generating query embeddings using an embedding model, a similarity metric such as cosine similarity is used to determine if the new query embeddings match the cached embeddings of past queries. This process involves setting a threshold for cosine similarity, which is a delicate balance.

In addition to privacy-preserving training of the embedding model, \tool also learns the optimal threshold ($\tau$) for cosine similarity. The range of $\tau$ is between 0 and 1. This threshold ($\tau$) dictates the level of similarity above which a cached query is considered relevant to the current user query. Setting the threshold too low could result in numerous false hits, leading to retrieving irrelevant queries from the cache. Conversely, a threshold set too high might cause many false misses, where relevant queries are not retrieved from the cache.  

During the client's local training, \tool determines this optimal threshold ($\tau$) from the client's feedback to the cache query response. Even after receiving a cached response, a user requests a response from the LLM, \tool considers it as a false positive and adjusts its threshold.  \tool varies the threshold $\tau$ to find the optimal threshold that optimizes the F-score of the cache (Section~\ref{sec:cosine-similarity-threshold-analysis}). By finding the optimal threshold, \tool effectively balances between true hits and true misses, therefore yielding improved accuracy in semantic similarity matching to return the response from the cache on duplicate queries.

\subsubsection{Aggregation} After client local training and finding the optimal threshold, each client sends updated weights of the global model (\gmodel) and optimal threshold ($\tau$) to the server (step 3 in Figure~\ref{fig:fl-training}). The server aggregates the updated weights from multiple users to form a new embedding model (\gmodel) using FedAvg~\cite{mcmahan2017communication} as shown in step 4 of Figure~\ref{fig:fl-training}. The server also computes the mean of the received optimal thresholds from the clients for a global optimal threshold ($\tau_{global}$). The benefit of finding $\tau_{global}$ is that when a new user joins the system, the user will not have queries to find its own optimal threshold. In such cases, the system can use $\tau_{global}$ as a starting point for semantic similarity. 

After finding the global optimal threshold and the global embedding model, the server then redistributes the updated embedding model to the users for the next round of FL training. This process is repeated over several FL rounds to improve the semantic matching accuracy (\ie lower false hits and false misses). 
After the completion of the FL training, each client will have access to a fine-tuned embedding model (\gmodel) to generate high-quality embeddings that can capture the complex semantics of a user query.

\begin{figure}[t]
    \centering
         \includegraphics[width=0.5\textwidth]{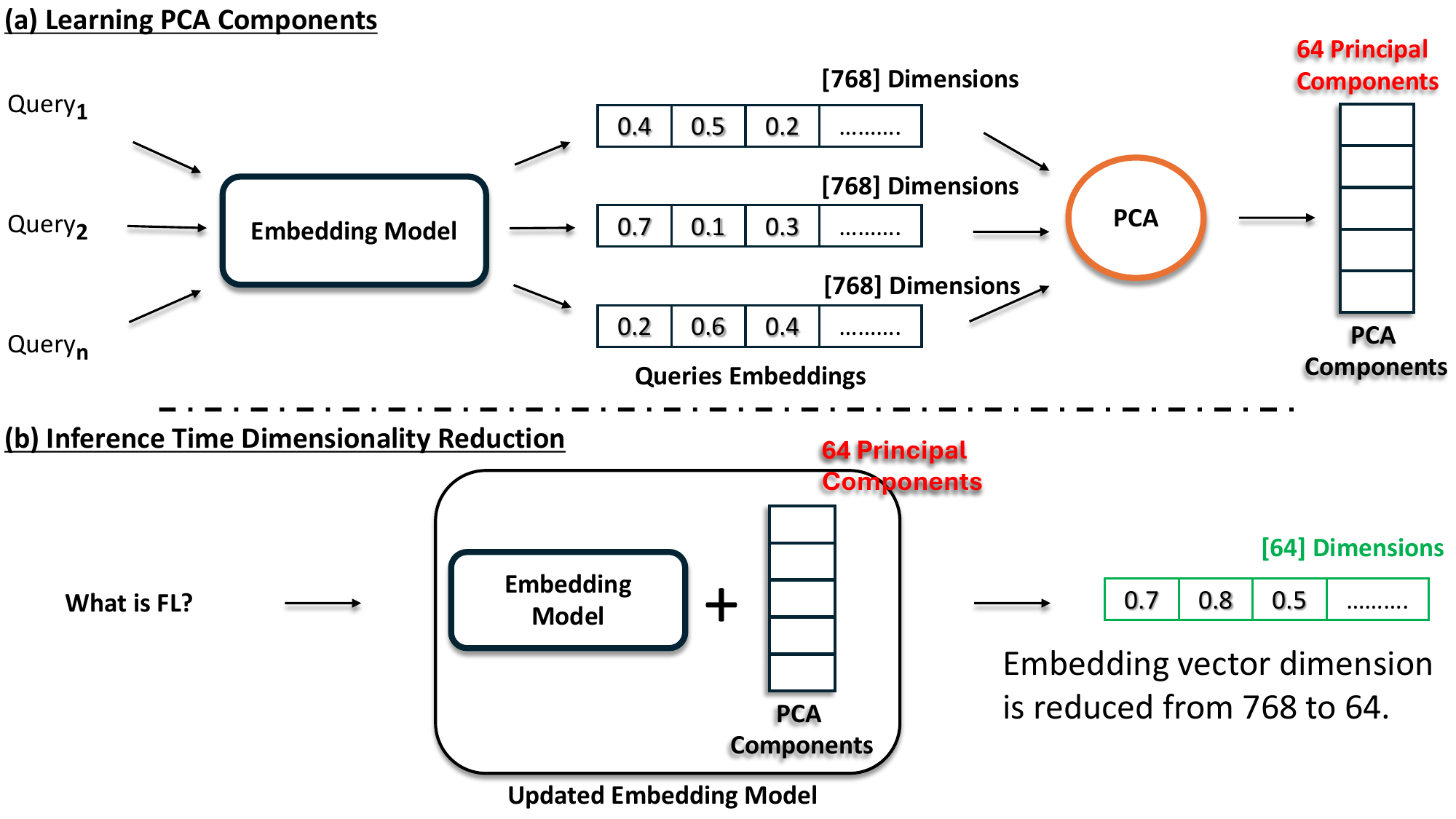}
          \vspace{-2ex}
          \caption{Embeddings Compression using PCA.}
          \label{fig:pca}
          \vspace{-2ex}
  \end{figure}

\subsubsection{Embeddings Compression using PCA}
The substantial size of the embedding vector (\eg Llama embeddings with a dimension of 4096) can lead to considerable overhead during the matching process of new query embeddings with cached queries embeddings. This is due to the search time being directly proportional to the dimensions of the embedding vector. Furthermore, high-dimensional embeddings demand more memory and storage. For example, the embeddings generated by Llama for a single query require approximately 32.05 KB of memory storage. Consequently, calculating the cosine similarity between two high-dimensional embeddings, specifically new query embeddings and each embedding in the cache, becomes computationally demanding and time-intensive. To improve the search time for identifying similar queries in the cache and to reduce the client's storage needs, it is essential to diminish the dimensionality of the embeddings while ensuring minimal impact on the \tool's performance.

PCA is a dimensionality reduction technique that is widely used to compress high-dimensional data into a lower-dimensional space ~\cite{pca1, pca2, pca3}, while still maintaining the most important information. First, \tool generates embeddings for all the users' queries using the embedding model. Next, \tool applies PCA to learn the principal components of all the queries embeddings generated in the previous step, as shown in Figure~\ref{fig:pca}-a. \tool integrates the learned principal components as an additional layer in the embedding model. This new layer will project the original embeddings onto the lower dimensional space, producing compressed embeddings (Figure~\ref{fig:pca}-b).

When a non-duplicate query is received, \tool uses the updated embedding model (with PCA layer) to generate the compressed embeddings (Figure~\ref{fig:pca}-b) for the new query and store the query, response, and the compressed embedding in the cache. Storing the compressed embeddings in the cache will significantly reduce the storage and memory overhead of the embeddings. Next, when a duplicate query is received, \tool uses the same embedding model with PCA components to generate the compressed embeddings for this duplicate query and find similar queries in the cache. Since the embeddings are compressed, the search time for finding similar queries in the cache will be significantly reduced.

\subsection{Cache Population and \tool Implementation}

Once the embedding model is trained within \tool on each user, it is deployed as depicted in Figure~\ref{fig:mean-cache-use-case}. Initially, when a new user starts using \tool, the local cache is vacant.  During these interactions, if a user query's response is not found in the cache, the request is forwarded to the LLM web service to retrieve the response, which is then inserted in the cache. If \tool finds semantically similar queries in the cache for any of the following queries from the user, it analyzes the context chain for every similar cached query and matches its embedding with the conversational history of the submitted query. If \tool finds a similar query with a similar context chain, the response is retrieved from the local cache and returned to the user. Otherwise, \tool forwards the query to the LLM web service to obtain the response. The query and its response and embeddings are then stored in the cache.

\tool is a python-based application that is built on the Flower FL  framework~\cite{beutel2020flower}. A user can submit LLM queries via this application to take advantage of the local cache.  The central server, which orchestrates the FL training, may reside on the LLM web service. We employ the Sbert~\cite{sbert-paper} library to train MPNet~\cite{song2020mpnet} and Albert~\cite{albert} on each client and to generate query embeddings. To efficiently execute a cosine similarity search between a query embedding and cached embeddings, we utilize Sbert's semantic search, which can handle up to 1 million entries in the cache. \tool cache storage is persistent and built using DiskCache~\cite{DiskCach54:online} library.

\subsection{Validating the Prevalence of Similar Queries}
\begin{figure}[t]
    \centering
    \includegraphics[width=0.5\textwidth]{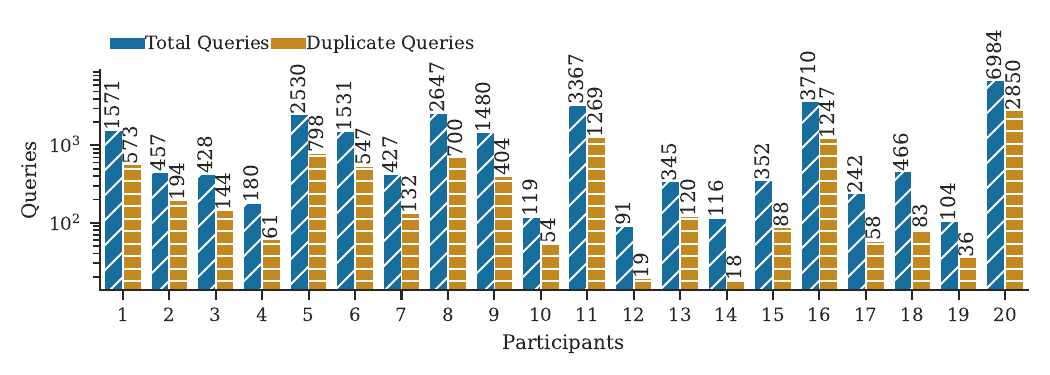}
    \caption{Analysis of real-world ChatGPT conversations reveals that, on average, 31\% of queries are similar to previously submitted queries.}
    \vspace{-3ex}
    \label{fig:survey}
\end{figure}

We conducted a privacy-preserving study with 20 ChatGPT users, analyzing over 27K queries. Our participants included university professors, developers, and graduate students who are regular ChatGPT users and experienced in running Python scripts. We provide these participants detailed instructions to download their queries and responses, run analysis scripts locally, and share aggregated results (\eg total and duplicate queries, field/profession). Individual queries are not shared, and thus, the conversations remain private. We find that about 31\% of user queries were similar to previous ones, suggesting a user-centric cache can reduce LLM inference costs (Figure~\ref{fig:survey}).  As this study was conducted in an academic setting, the ratio of repeated queries may vary in other contexts.


\section{Evaluation}
\label{sec:evaluation}

Our evaluation answers the following research questions.

\begin{itemize}
    \item How does \tool perform in comparison to baseline in terms of performance metrics (\cref{sec:eval_comp_baseline})?
    \item How accurately does \tool retrieve contextual queries from the cache (\cref{sec:contextual-queries})?  
    \item Is it possible to reduce the embedding dimension to save storage space and accelerate semantic search while outperforming the baseline (\cref{sec:storage_eval})? 
    \item Is it possible to train an embedding model in a privacy-preserving manner without the centralized data (\cref{sec:eval_privacy})?
    \item What effect does the cosine similarity threshold have on the performance of \tool (\cref{sec:cosine-similarity-threshold-analysis})?
    \item GPTCache~\cite{gptcache} suggests using Llama 2 to generate embeddings to improve semantic matching. Is it feasible to use Llama 2 to compute embeddings at the user side for semantic matching (\cref{sec:Llama-2-for-embedding-computation})?
\end{itemize}

\subsection{Evaluation Settings}

We conduct evaluations of \tool against the baseline~\cite{gptcache} to demonstrate that \tool achieves optimal performance while preserving user-privacy (\ie without storing the user queries at the server). For a fair comparison between \tool and baseline, we employ the optimal configuration as described in the GPTCache study~\cite{gptcache}. This configuration utilizes Albert~\cite{albert} and sets the cosine similarity threshold at 0.7 to determine the cache hit or miss. 

\begin{table}[t]
    \centering
    \scalebox{1}{
    \begin{tabular}{|l|l|l|l|l|l|}
    \hline
    \textbf{Metrics} & \multicolumn{3}{|c|}{\textbf{Standalone Queries}} & \multicolumn{2}{|c|}{\textbf{Contextual Queries}} \\ \hline
    & \textbf{GPT-} & \textbf{Mean-} & \textbf{Mean-} & \textbf{GPT-} & \textbf{\tool} \\
    & \textbf{Cache} & \textbf{Cache} & \textbf{Cache} & \textbf{Cache} &  \\
    &  & (MPNet) & (Albert) & & \\ \hline
    F score & 0.56 & 0.73 & 0.68 & 0.67 & {\bf 0.93} \\ \hline
    Precision & 0.52 & 0.72 & 0.66 & 0.66 & {\bf 0.98} \\ \hline
    Recall &  0.85 & 0.78 & 0.77 & 0.71 &  0.79 \\ \hline
    Accuracy & 0.72 & 0.85 & 0.81 & 0.61 & {\bf 0.86} \\ \hline
    \end{tabular}}
    \caption{\tool outperforms GPTCache (baseline) on both standalone and contextual queries. }
    \label{table:tool-vs-gptcache}
\end{table}

\begin{figure*}[t]
    \centering
    \resizebox{0.8\textwidth}{!}
    {\input{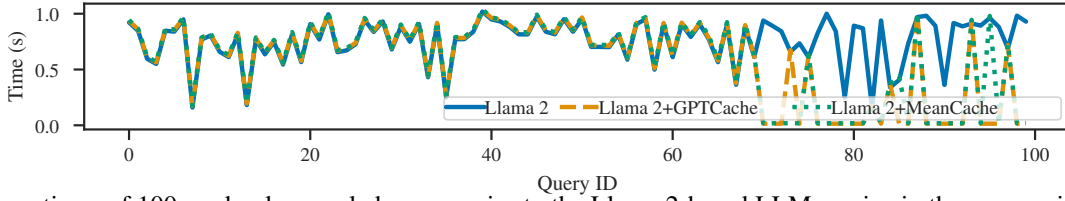}}
    \vspace{-3ex}
    \caption{Response times of 100 randomly sampled user queries to the Llama 2-based LLM service in three scenarios: without any semantic cache, with GPTCache, and with \tool. The integration of a semantic cache does not add significant overhead to non-duplicate queries, meaning it does not impede the performance of the LLM-based service. Moreover, it significantly reduces the average response times for duplicate queries (70-99) by serving them from the local cache.}
    \label{fig:Llama 2-end2end-time-cache-overhead}
\end{figure*}

\begin{figure*}[t]
    \centering
    \resizebox{0.8\textwidth}{!}
    {\input{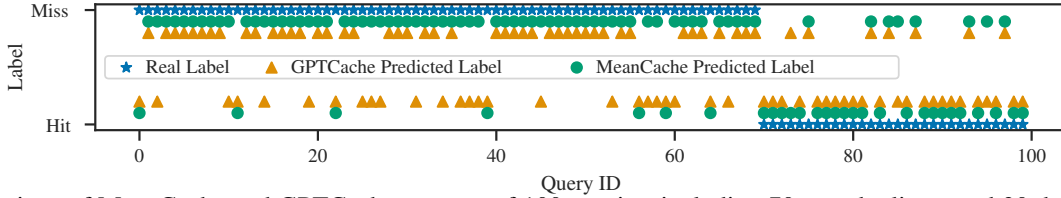}}
    \vspace{-3ex}
    \caption{Comparison of \tool and GPTCache on a set of 100 queries, including 70 non-duplicate and 30 duplicate queries, sent to the Llama 2-based LLM service. Queries 0 to 69 are non-duplicate (\ie~{\em real label is miss}), and GPTCache produces significantly higher false hits on these unique queries compared to \tool.}
    \label{fig:llama2_end2end_hitmiss}
\end{figure*}

\begin{figure}[t]
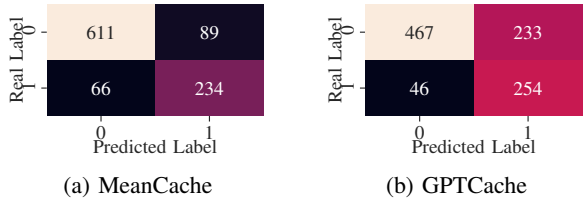

    \centering
    \begin{subfigure}{0.4\columnwidth}
        \centering
        \resizebox{\columnwidth}{!}
        {\input{./graphs/pgf/final_llama2_times_end_to_end_with_dup_query_percent_0.3hit_miss_fedgptcache_confusion_matrix.pgf}}
        \caption{\tool}
        \label{fig:cm_fedgptcache}
    \end{subfigure}
    \hfil
    \begin{subfigure}{0.4\columnwidth}
        \centering
        \resizebox{\columnwidth}{!}
        {\input{./graphs/pgf/final_llama2_times_end_to_end_with_dup_query_percent_0.3hit_miss_gptcache_confusion_matrix.pgf}}
        \caption{GPTCache}
        \label{fig:cm_gptcache}
    \end{subfigure}

    \caption{Confusion matrices for \tool and GPTCache on 1000 queries comprising 700 unique and 300 duplicate queries. Among the 700 unique queries, \tool produces only 89 false hits, while GPTCache generates 233  false hits.}

    \label{fig:cm}
\end{figure}

\subsubsection{Transformer Models and Datasets} \label{sec:evaluation-dataset}
For extensive evaluations of \tool, we utilize the Llama 2~\cite{llama}, MPNet~\cite{song2020mpnet}, and Albert~\cite{albert} transformer models to generate embeddings.

We evaluate \tool using the GPTCache dataset. The dataset is partitioned into training, testing, and validation subsets. The training and validation datasets are randomly distributed among the clients, each receiving non-overlapping data points. During local training, each client utilizes its training dataset to update its local embedding model and employs the validation dataset to determine the optimal threshold for cosine similarity (Section~\ref{sec:cosine-similarity-threshold-analysis}). The testing dataset, located at the central server, facilitates a fair comparison between \tool and GPTCache~\cite{gptcache}. Since there does not exist any dataset of contextual queries, we generate a synthetic dataset using GPT-4 consisting of 450 queries, including duplicates, non-duplicates, and contextual queries, to evaluate \tool performance on contextual queries.

\subsubsection{Experimental Setup} The experiments are conducted on a high-performance computing cluster, equipped with 128 cores, 504 GB of memory, and four A100 Nvidia GPUs, each with 80 GB of memory. We utilize the Flower FL~\cite{beutel2020flower} library to simulate a FL setup. Additionally, the SBERT~\cite{sbert-paper} library is employed to train the embedding model on each client. The number of clients participating in FL training are 20. The number of clients is restricted due to the limited size of the GPTCache dataset, which is inadequate for distribution among hundreds of clients. However, we believe \tool results are not influenced by the number of clients, and the evaluation setup of \tool is consistent with the evaluation standard in FL~\cite{avdiukhin2021federated, wang2020tackling}.

\subsubsection{Evaluation Metrics}
In caching systems, the efficacy has traditionally been gauged by cache hit-and-miss rates. A cache hit implies the data or query is retrieved from the cache, whereas a cache miss indicates the opposite. Semantic caching introduces a nuanced classification: true and false hits, alongside true and false misses. A {\em true hit} signifies a correct match between a query and a similar cached query, whereas a {\em false hit} is an incorrect match with a non-similar cached query. A {\em true miss} signifies when a query does not have a similar cached query, whereas a {\em false miss} happens when a query has a similar cached query but is not returned from the cache. Thus, traditional hit/miss metrics are potentially misleading in semantic caches. For example, a query might incorrectly match with an irrelevant cached query (deemed a hit traditionally) due to semantic matching. We adopt precision, recall, F score, and accuracy for a comprehensive evaluation of \tool against baseline. These metrics are defined as follows:

\noindent{\textbf{Precision.}} The ratio of true positive hits to all positive hits (including both true positives and false positives). In semantic caching, this measures how many of the queries matched to a cached query are correctly matched. $\text{Precision} = \frac{TP}{TP + FP}$ where $TP$ represents true positives (true hits) and $FP$ represents false positives (false hits).

\noindent{\textbf{Recall.}} The ratio of true positive hits to all relevant items (including both true positives and false negatives). In semantic caching, this assesses the proportion of correctly matched queries out of all queries that should have been matched to a cached query. $\text{Recall} = \frac{TP}{TP + FN}$ where $FN$ represents false negatives (false misses).

\noindent{\textbf{F\(_\beta\) Score.}} A weighted harmonic mean of precision and recall, balancing the two based on the value of \(\beta\). \(\beta > 1\) gives more weight to recall, while \(\beta < 1\) emphasizes precision.
\[
    F_\beta = (1 + \beta^2) \cdot \frac{\text{Precision} \times \text{Recall}}{(\beta^2 \times \text{Precision}) + \text{Recall}}
\]

\noindent{\textbf{Accuracy.}} The ratio of correctly identified queries (both true hits and true misses) to all queries. $\text{Accuracy} = \frac{TP + TN}{TP + TN + FP + FN}$ where $TN$ represents true negatives (true misses).

\subsection{\tool Comparison with Baseline}
\label{sec:eval_comp_baseline}
We evaluate \tool against GPTCache to assess improvements in precision, recall, and F score. 
MeanCache FL model training is discussed in Section~\ref{sec:eval_privacy}, and the optimal threshold is covered in Section~\ref{sec:cosine-similarity-threshold-analysis}.
 
 We select a sample of 1000 queries, 30\% of which are repeated queries (\ie 300 queries are repeated), and load these 1000 queries as cached queries. Note that repeated queries are usually fewer than non-repeated queries. Thus, we use 30\% as repeated queries, a similar percentage previously observed for web services~\cite{markatos2001caching}.

Initially, we send a new set of one thousand queries to Llama 2 (i.e., without any semantic cache) to establish a baseline for response times. We limit responses to 50 tokens to reflect practical response sizes, although actual sizes can be much larger. Note that \tool's performance is not dependent on the response as it only matches the queries. Next, we send these queries to Llama 2 based local LLM service with \tool and GPTCache to measure the response times and performance metrics (\eg precision, recall, F-score), respectively. An analysis of a random subset of 100 queries (70 non-duplicate and 30 duplicate) from the 1000 queries shows the impact of caching on response times in Figure~\ref{fig:Llama 2-end2end-time-cache-overhead} and cache hit and miss rates in Figure~\ref{fig:llama2_end2end_hitmiss}. 
While \tool is evaluated on 1,000 standalone and 450 contextual queries (Table~\ref{table:tool-vs-gptcache}), we use this smaller subset of 100 queries solely for visualization purposes, as displaying all queries would create cluttered figures. 
The y-axis in Figure~\ref{fig:Llama 2-end2end-time-cache-overhead} shows the response time of each query without any cache (Llama 2), with GPTCache (Llama 2 + GPTCache), and with \tool (Llama 2 + \tool). In Figure~\ref{fig:llama2_end2end_hitmiss}, x-axis represents the query and the y-axis represents each semantic cache hit and miss alongside the real label. Figure~\ref{fig:Llama 2-end2end-time-cache-overhead} demonstrates that implementing a semantic cache does not impede the performance (queries ranging from 0 to 69) and improves the user experience as response times reduce on duplicate queries (queries 70 to 99).

However, Figure~\ref{fig:llama2_end2end_hitmiss} shows that GPTCache produces significant false hits on non-duplicate queries (queries 0 to 69) compared to \tool. Each false hit means the user receives an incorrect cached response, requiring them to resend the query to the LLM service, leading to a poor user experience. To prioritize precision, we adjust the beta value in the F score to 0.5, valuing precision twice as highly as recall to ensure user satisfaction by avoiding false positives. This decision is driven by the need to minimize user inconvenience caused by incorrect cache hits. A false miss (low recall) does not require user intervention as the false miss query will be automatically routed to the LLM. Thus, precision in semantic caching is more important than recall.

Table \ref{table:tool-vs-gptcache} and the confusion matrices in Figure~\ref{fig:cm} highlight \tool's superior performance over GPTCache on the 1000 user queries. Notably, \tool with MPNet achieves a precision of 0.72, significantly surpassing GPTCache's 0.52. This superiority is evident in the lower false positive rates (i.e., false hits) shown in Figure~\ref{fig:cm_fedgptcache}. The number of false hits for \tool is 89 (Figure~\ref{fig:cm_fedgptcache}), while GPTCache has 233 false hits, as depicted in Figure~\ref{fig:cm_gptcache}. In practical terms, this means that with GPTCache, the end user has to manually resend 233 queries to the LLM service to get the correct responses, compared to only 89 queries with \tool. While GPTCache's recall is higher than \tool's, as we discussed earlier, precision is significantly more important than recall, and \tool outperforms GPTCache in this regard. Overall, the F-score of \tool with MPNet is 0.73 and 0.68 with the Albert embedding model, both of which outperform GPTCache's F-score of 0.56.

\begin{tcolorbox}[colback=white, colframe=black,  left=0pt, right=0pt, top=0pt, bottom=0pt]
    \textbf{\textit{Summary.}} In an end-to-end deployment, \tool significantly outperforms GPTCache. It demonstrates a 17\% higher F score and a 20\% increase in precision in optimal configuration. The substantial reduction in false cache hits enhances the end-user experience.
\end{tcolorbox}

\subsection{Contextual Queries}
\label{sec:contextual-queries}

\begin{figure}[t]
    \centering
    \begin{subfigure}{\columnwidth}
    \centering
    \includegraphics[width=1\textwidth]{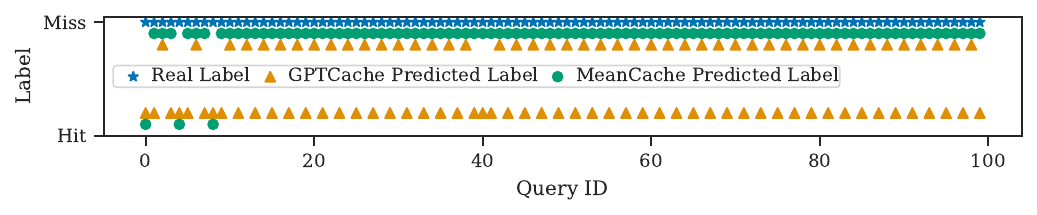}

    \caption{Ideally, all 100 queries should result in misses. However, GPTCache incorrectly produces 54 \emph{false hits}, while \tool yields only 3.}
    \label{fig:only_miss_contextual_queries}
            
    \end{subfigure}
    \vfil
    \begin{subfigure}{\columnwidth}
    \centering
    \includegraphics[width=1\textwidth]{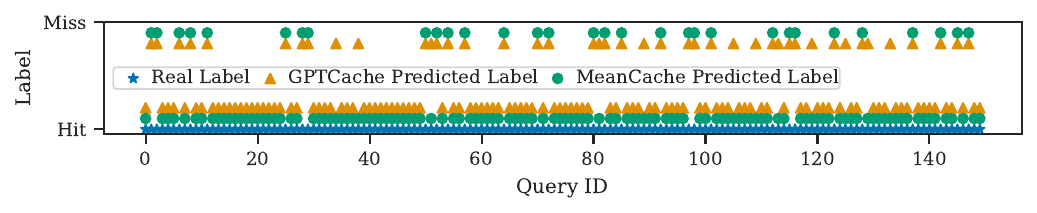}
    \caption{\tool yields 8\% more \emph{true hits} than the baseline.}
    \label{fig:only_hit_contextual_queries}
    \end{subfigure}

    \caption{Performance on Contextual Queries: \tool vs. Baseline. (a) reports \tool's fewer false hits 3 vs. 54 of GPTCache. (b) reports higher true hits by \tool.}

    \label{fig:predictions-hit-miss-graphs}
\end{figure}

\begin{figure}[t]
    \centering
    \begin{subfigure}{0.49\columnwidth}
        \centering
    \includegraphics[width=1\textwidth]{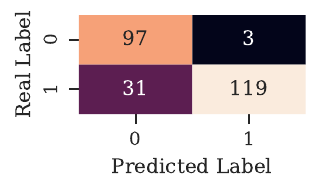}

    \caption{\tool}
    \end{subfigure}
    \hfil
    \begin{subfigure}{0.49\columnwidth}
    \centering

    \includegraphics[width=1\textwidth]{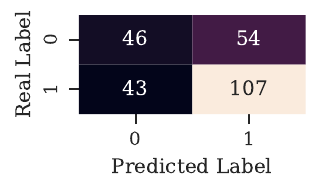}
    
    \caption{GPTCache (baseline)}
    \end{subfigure}

    \caption{For contextual queries, \tool reports three false hits, compared to 54 false hits by GPTCache.}
    \vspace{-2ex}
    \label{fig:contextual-cm}
\end{figure}

Section~\ref{sec:background} describes contextual queries. GPTCache, lacking the capability to detect contextual queries, incorrectly identifies such queries as cache hits, resulting in inaccurate responses. \tool addresses this limitation by verifying the context chain (Algorithm~\ref{alg:tool}) of semantically matched queries. This verification ensures that contextual queries (\eg Q4 in Section~\ref{sec:background}) correctly yield a \emph{true cache miss}, thus prompting an appropriate request to the LLM service.

On the dataset of 450 contextual queries (see Section~\ref{sec:evaluation-dataset}),  first, we populate the cache (\tool and baseline) with 200 queries (100 standalone and 100 contextual queries of the standalone queries). Next, we send 150 duplicate queries  (75 standalone + 75 contextual) and 100 non-duplicate queries (a total of 250 queries) to the cache-enabled LLM. Figure~\ref{fig:predictions-hit-miss-graphs} shows the true label (whether the query should be returned from the cache or not) and the corresponding GPTCache and MeanCache performance (predicted label). Note that in Figure~\ref{fig:only_miss_contextual_queries}, all the queries should be answered by the LLM; in other words, there should be no hits. 
 However, GPTCache has 54 false hits, while \tool has only three false hits. This is also shown in the confusion matrix in Figure~\ref{fig:contextual-cm}. Table \ref{table:tool-vs-gptcache} (Column-3) summarizes the comparative results. \tool outperforms GPTCache by over 25\% in both F-Score and accuracy. Additionally, \tool achieves 32\% higher precision compared to the baseline.

\begin{tcolorbox}[colback=white, colframe=black,  left=0pt, right=0pt, top=0pt, bottom=0pt]
    \textbf{\textit{Summary.}} GPTCache's low performance stems from its inability to consider contextual information, leading to high false hit rates. \tool demonstrates superior handling of contextual queries, with a 25\% improvement in accuracy over the baseline.
\end{tcolorbox}

\begin{figure*}[t]
    \centering
    \resizebox{0.90\textwidth}{!}
    {\input{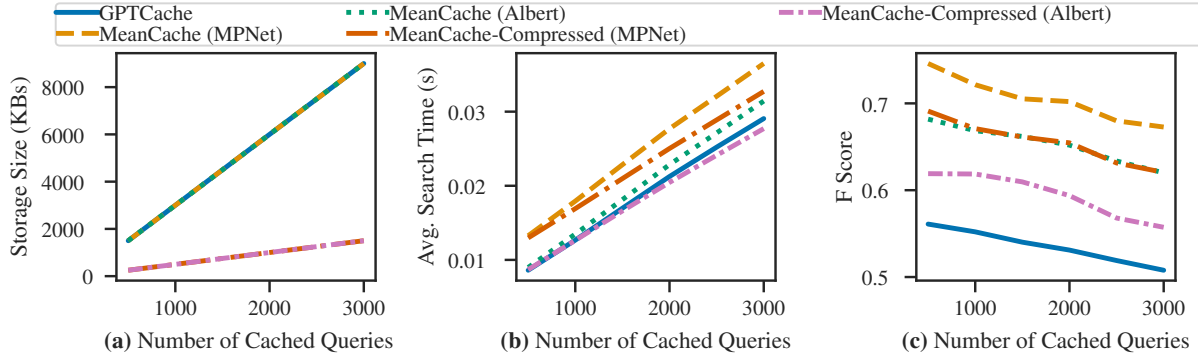}}
    \caption{(a)~\tool's embedding compression reduces storage  by 83\% compared to GPTCache. (b) \tool's semantic matching speed is 11\% faster with compression enabled, while still outperforming GPTCache. (c) \tool's F score is slightly lower with compression enabled, but it still outperforms GPTCache.}
    \vspace{-3ex}
    \label{fig:embedding-dimension-reduction}
\end{figure*}

\subsection{Embedding Compression and Impact on Storage Space}
\label{sec:storage_eval}
Clients often face storage limitations compared to web servers. Storing embeddings in the local cache on the user side for semantic search demands memory storage. Various models yield embeddings with differing vector sizes; for example, the MPNet and Albert models produce an output embedding vector of 768 dimensions, whereas the Llama 2 model's embeddings dimension size is 4096. The embedding vector size also affects semantic search speeds, where smaller vectors could enhance speed and lower resource demands.

Figure~\ref{fig:embedding-dimension-reduction} illustrates the effects of \tool dimension reduction utility on the storage, semantic matching speed (overhead), and \tool's performance (F score). The x-axis indicates the number of queries stored in the cache, while the y-axis shows storage size, average search time, and the F score in respective graphs. \tool-Compressed (MPNet) and \tool-Compressed (Albert) represent instances where \tool decreases the embedding dimensions from 768 to 64 by employing the compression, as detailed in \tool design (Section~\ref{sec:tool-design}).

\begin{figure}
        \centering
        \resizebox{0.36\textwidth}{!}
        {\input{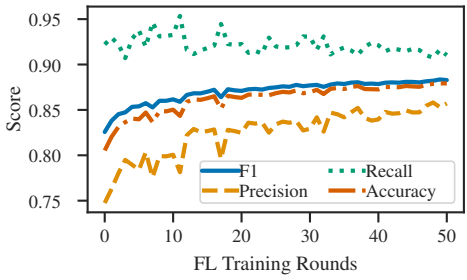}}
        \caption{\emph{MPNet's} FL training helps generate high-quality embeddings.}
        \label{fig:training_mpnet}
    \end{figure}

\begin{figure}
        \centering
        \resizebox{0.36\textwidth}{!}
        {\input{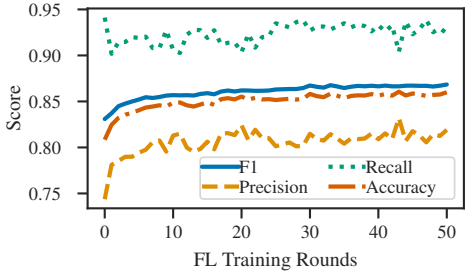}}
        \caption{FL training boosts \tool's query matching precision with \emph{Albert}.}
        \label{fig:training_albert}
    \end{figure}

Figure~\ref{fig:embedding-dimension-reduction} demonstrates that increased stored queries linearly raise storage needs. Yet \tool with compression enabled drastically lowers storage needs by 83\% compared to GPTCache. Figure~\ref{fig:embedding-dimension-reduction} also indicates that compression decreases the average search time, with \tool enabled compression approximately 11\% faster. Moreover, despite a slight decrease in F score with compression enabled, \tool still surpasses GPTCache. Furthermore, given the evidence from  Section~\ref{sec:eval_privacy} (Figures~\ref{fig:training_mpnet} and \ref{fig:training_albert}) that MPNet produces more precise embeddings, and it is also clear from Figure~\ref{fig:embedding-dimension-reduction} that MPNet's embeddings are particularly resilient to compression and excel in semantic matching.

\begin{tcolorbox}[colback=white, colframe=black,  left=0pt, right=0pt, top=0pt, bottom=0pt]
    \textbf{\textit{Summary.}} The application of embedding compression optimization in \tool offers substantial benefits, including an 83\% savings in storage and an 11\% faster search process, while still outperforming the established baseline (GPTCache).
\end{tcolorbox}

    

\subsection{Privacy Preserving Embeddings Model Training}
\label{sec:eval_privacy}
Storing clients queries on the server side presents a potential privacy risk. To address this, each client can retain its local data on its own device. The ensuing challenge is how to train an embedding model that can also utilize the distributed data from all clients. FL is recognized for training neural networks in a privacy-preserving manner. As such, \tool employs FL to train and fine-tune an embedding model, thereby preserving privacy and leveraging the dataset residing on the client's side. In this section, our objective is to evaluate whether FL training can progressively enhance the embedding model to generate high-quality embeddings for user queries. To simulate this scenario, we distribute the training dataset among 20 clients. In each round, we sample 4 clients, conducting a total of 50 FL training rounds. Each client trains its embedding model for 6 epochs, operating on a dedicated A100 GPU.  We conduct two experiments with the Albert and MPNet models. The batch size is set to 256 for the Albert model, and for MPNet, it is set to 128 during the local training by participating clients.

Figures~\ref{fig:training_mpnet} and~\ref{fig:training_albert} depict the performance of \tool as FL training progresses. The x-axis represents the training round, while the y-axis shows the performance metrics such as F-score, precision, recall, and accuracy of the global model (\gmodel). As illustrated in Figure~\ref{fig:training_mpnet}, the F-score for MPNet increases from 0.82 to 0.88, and for Albert, it rises from 0.83 to 0.86, as shown in Figure~\ref{fig:training_albert}. Similarly, precision for MPNet significantly increases from 0.74 to 0.85, as depicted in Figure \ref{fig:training_mpnet}, and for Albert, it increases from 0.74 to 0.81, as demonstrated in Figure \ref{fig:training_albert}. Given that MPNet is a more robust transformer architecture compared to Albert, it is also observed during our training that MPNet outperforms Albert, exhibiting superior learning in FL settings.

\begin{tcolorbox}[colback=white, colframe=black,  left=0pt, right=0pt, top=0pt, bottom=0pt]
    \textbf{\textit{Summary.}} FL training increases 11\% precision of \tool for MPNet and a 7\% increase for Albert. The performance of the embedding model to generate high-quality embeddings can improve in a privacy-preserving manner using FL training.
\end{tcolorbox}

\begin{figure}
        \centering
        \resizebox{0.36\textwidth}{!}
        {\input{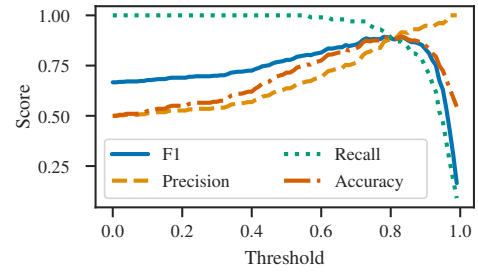}}
        \caption{\tool optimizes MPNet's threshold.}
        \label{fig:threshold_variation_mpnet}
    \end{figure}
    
    \begin{figure}
        \centering
        \resizebox{0.36\textwidth}{!}
        {\input{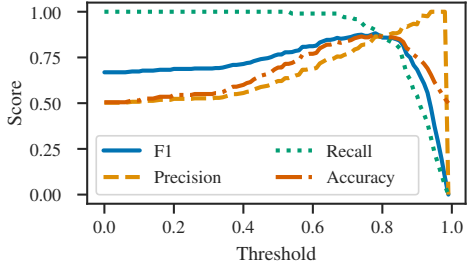}}
        \caption{\tool identifies an optimal threshold of 0.78 for Albert.}
        \label{fig:threshold_variation_albert}
    \end{figure}

\subsection{Cosine Similarity Threshold Impact on Semantic Matching}
\label{sec:cosine-similarity-threshold-analysis}

Semantic matching for a new user query begins by generating the embeddings of the user's query ($E_q$) using the embedding model. The cosine similarity ($\theta$) is computed with the cached embeddings. If the cosine similarity $\theta$ exceeds the threshold $\tau$, the cache is hit and the response to the user query is returned from the cache. Therefore, the cosine similarity threshold $\tau$ is crucial in determining the similarity between a user query and cached entries. A low threshold value of $\tau$ can lead to false hits (incorrect matches), while a high threshold might overlook the appropriate matches (\ie false cache misses or false negatives). 

To illustrate this, \tool varies the threshold $\tau$ from 0 to 9 and evaluates the performance metrics F-score, precision, recall, and accuracy with an equal distribution of duplicate and non-duplicate queries from the validation data to avoid bias. Figures \ref{fig:threshold_variation_mpnet} and \ref{fig:threshold_variation_albert} show how the cosine similarity threshold ($\tau$) affects \tool's performance. The x-axis represents the threshold $\tau$ values, and the y-axis denotes the performance metrics. For instance, at a 0.3 threshold, \tool's semantic matching accuracy using MPNet is 57\%, with a precision of 54\% as shown in Figure~\ref{fig:threshold_variation_mpnet}. Similarly, with Albert at the same threshold, the accuracy is 55\%, and the precision is 53\% (Figure~\ref{fig:threshold_variation_albert}). Precision typically improves with an increase in threshold. However, beyond a certain point, increasing the threshold $\tau$ leads to a decline in F score, accuracy, and recall.

For MPNet, the optimal threshold $\tau$ is identified at 0.83, achieving an F1 score of 0.89, precision of 0.92, and accuracy of 0.90 (Figure~\ref{fig:threshold_variation_mpnet}). For Albert, the optimal threshold is 0.78, with an F1 score of 0.88.

\begin{tcolorbox}[colback=white, colframe=black,  left=0pt, right=0pt, top=0pt, bottom=0pt]
    \textbf{\textit{Summary.}}  GPTCache's suggested threshold of 0.7 will result in suboptimal performance during semantic matching. The optimal threshold  $\tau$ values varies with the embedding model. \tool optimally adjusts the threshold based on user data, outperforming GPTCache's suggested threshold by 16\% in precision and 4\% in F score for MPNet, and by 10\% in precision and 2\% in F score for Albert. 
\end{tcolorbox}

\subsection{Infeasibility of Embedding Generation with Llama 2}
\label{sec:Llama-2-for-embedding-computation}

GPTCache~\cite{gptcache} recommend using Llama for generating embeddings to enhance GPTCache's performance. However, Llama 2's embedding computation is expensive in terms of inference time, requires substantial storage, and incurs considerable overhead during semantic searches. For example, Llama 2 with its 7 billion parameters, demands 30 GB of memory~\cite{SizingGu24:online}, whereas Albert and MPNet require only 43 MB and 420 MB, respectively.

\begin{figure}[t]
    \centering
    \resizebox{0.36\textwidth}{!}
    {\input{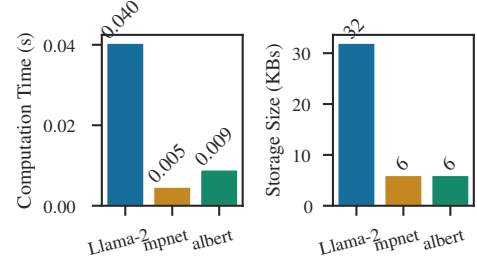}}
    \caption{Llama 2 takes significantly longer to compute embeddings and requires substantially more storage space than Albert and MPNet.}
    \label{fig:inference_storage_overhead}
\end{figure}

\begin{figure}[t]
    \centering
    \resizebox{0.36\textwidth}{!}
    {\input{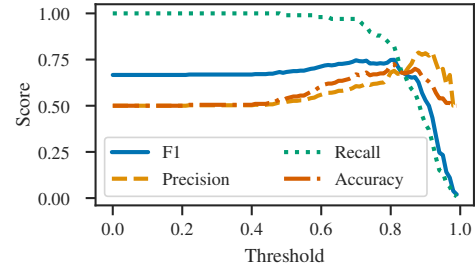}}
    \caption{Llama 2 does not perform well in semantic matching, even at optimal thresholds.}
    \vspace{-2ex}
    \label{fig:llama2_threshold}
\end{figure}

To highlight the impracticality to generate query embeddings with Llama 2, we compare the embedding computation time, embedding storage space requirement of Llama 2, Albert and MPNet transformer models. Figure~\ref{fig:inference_storage_overhead} shows the average time to compute the embeddings of a single query and storage requirements for the embeddings. Figure~\ref{fig:inference_storage_overhead} shows that the average embedding computation time of Llama 2 is 0.04 seconds, while for Albert and MPNet the average computation time is 0.005 and 0.009 seconds respectively. Single query embeddings generated from Llama 2 takes approximately 32 KBs of space and embeddings generated by both MPNet and Albert only take 6 KBs, as shown in Figure~\ref{fig:inference_storage_overhead}.

Furthermore, we evaluate the performance of the Llama 2 on embedding generation and semantic matching. Figure~\ref{fig:llama2_threshold} shows that the performance of Llama 2 with different cosine similarity threshold ($\tau$) and corresponding performance metric. We can see that the performance of Llama 2 is not good even with the optimal cosine similarity threshold, as also noted by researchers \cite{UserEmbe50:online}. The maximum F1 score achieved by Llama 2 is 0.75 which is quite low when compared with the optimal thresholds scores from the Figures~\ref{fig:threshold_variation_mpnet} and Figure~\ref{fig:threshold_variation_albert}.

Smaller models tailored for specific tasks often surpass larger models in efficiency~\cite{ouyang2022training, penedo2024refinedweb, du2024compositional}. Thus, diverging from GPTCache's approach, we advocate for adopting smaller yet efficient embedding models for semantic caching. These models not only ensure optimal performance but also minimize the semantic cache's overhead on users, featuring lower inference demands and reduced output embedding sizes, thereby facilitating deployment on edge devices.

\begin{tcolorbox}[colback=white, colframe=black,  left=0pt, right=0pt, top=0pt, bottom=0pt]
    \textbf{\textit{Summary.}}  Llama 2  is not viable for generating embeddings. Future enhancements might improve its performance to generate embeddings, but the computational demands, semantic search duration, and storage requirements will likely remain elevated.
\end{tcolorbox}

\section{Related Work}
\label{sec:related-work}

Several caching systems are proposed to optimize the performance. Study~\cite{saraiva2001rank} suggests a two-tier dynamic caching architecture for web search engines to enhance response times in hierarchical systems. Utilizing LRU eviction at both levels, they demonstrate how the second-tier cache can significantly lower disk traffic and boost throughput. Researchers in~\cite{baeza2003three} propose a three-level index organization and~\cite{long2005three} propose a three-tier caching. Another study~\cite{xie2002locality} examined two real search engine datasets to explore query locality, aiming to develop a caching strategy based on this concept. Their analysis centered on query frequency and distribution, assessing the feasibility of caching at various levels, such as server, proxy, and client side. A novel caching technique called Probability Driven Cache (PDC) is proposed in~\cite{lempel2003predictive} to optimize the performance of search engines by using the probability of query repetition to decide whether to cache the query. PDC uses the probability of a query to be repeated to decide whether to cache the query or not. A different approach is presented in~\cite{fagni2006boosting}, which proposes the Static Dynamic Cache (SDC) to exploit temporal and spatial locality present in the query stream, avoiding redundant processing and saving computational resources. Efficient caching designs for web search engines are explored in~\cite{baeza2007impact}, where static and dynamic caching strategies are compared, weighing the benefits of caching query results against posting lists. Challenges in large-scale search engines, which process thousands of queries per second across vast document collections, are examined in~\cite{zhang2008performance}, focusing on index compression, caching optimizations, and evaluating various inverted list compression algorithms alongside caching policies such as LRU and LFU.

All of these studies focus on caching systems designed for traditional search engines that process keyword queries (i.e., keyword matching) and return a list of links as a response. However, when applied to LLM-based web servers or APIs, these caching systems do not provide a single concise response and may yield many false results. Moreover, such caching techniques fail to capture the semantic similarity among repeated queries, leading to a significantly low hit rate. Server-side caching for services based on LLMs is proposed in~\cite{zhu2023towards} and~\cite{gptcache}, aiming to reduce the massive computational cost of LLMs. In particular, the approach in~\cite{zhu2023towards} checks if a new query is semantically similar to any existing queries in the cache. If a match is found, the cached response is returned; otherwise, a model multiplexer selects the most suitable LLM.

While these techniques can handle semantic similarity among queries and provide a single concise response, they raise privacy concerns as user queries are stored on external servers. Additionally, these techniques are static and unable to adapt to individual user behavior. Users may still be charged for the query, even if it is served from the cache. Therefore, a user-centered semantic cache that operates on the user side is needed, providing benefits in terms of privacy, cost, and latency. This cache should be able to detect semantic similarity among queries and adapt to each user's behavior while preserving privacy. \tool offers these benefits without compromising user privacy.

\section{Conclusion}
\label{sec:conclusion}
\tool introduces the first user-centric semantic cache designed for LLM-based web services, such as ChatGPT. In \tool, clients collaboratively train a global embedding model using FL on their local data, ensuring user privacy. After aggregation, the global model produces high-quality embeddings for effective semantic matching. When a new query from the user matches a previous one, \tool semantically compares it with the user's local cache and retrieves the most relevant results. This approach reduces the computational cost of LLM services, enhances bandwidth and latency, and conserves the user's query quota. Even with compressed embeddings that save 83\% of storage space, \tool outperforms existing baseline. With its distributed cache design, \tool offers a solution to reduce up to one-third of LLM query inference costs for semantically similar queries on the user side. 

\noindent \textbf{Acknowledgment.}
We thank anonymous reviewers for providing valuable and constructive feedback to help improve the quality of this work.
This work was supported by Amazon - Virginia Tech Initiative in Efficient and Robust Machine Learning. We also thank the Advanced Research Computing Center at Virginia Tech and the Flower FL framework for their support in building and evaluating this work.

\bibliographystyle{IEEEtran}
\bibliography{main}

\end{document}